\documentclass[12pt, a4paper]{article}

\usepackage[utf8]{inputenc}
\usepackage[english]{babel}
\usepackage{graphicx}
\usepackage{amsmath}
\usepackage{amssymb}
\usepackage{geometry}
\usepackage{hyperref}
\usepackage{natbib}
\usepackage{authblk}
\usepackage{setspace}
\usepackage{booktabs}
\usepackage{subcaption}
\usepackage{pifont}
\usepackage{todonotes}
\usepackage{enumitem}
\usepackage{csquotes}
\usepackage{listings}

\usepackage{parskip}  % Don't indent paragraphs. Sets \parindent to 0 and adds reasonable space between paragraphs.

\hypersetup{
    colorlinks=true,
    linkcolor=blue,
    filecolor=magenta,
    urlcolor=cyan,
    citecolor=red,
}

\title{Limits of Spatial Imagery Reasoning in Frontier LLM Models}
\author[1]{Sergio Y. Hayashi}
\author[1]{Nina S. T. Hirata}
\affil[1]{Institute of Mathematics and Statistics -- University of S\~ao Paulo, Brazil}
\date{\today}

\begin{document}

\maketitle

% --- Abstract ---
\begin{abstract}
\noindent
Large Language Models (LLMs) have demonstrated impressive reasoning capabilities, yet they struggle with spatial tasks that require mental simulation, such as mental rotation. This paper investigates whether equipping an LLM with an external ``Imagery Module''---a tool capable of rendering and rotating 3D models---can bridge this gap, functioning as a ``cognitive prosthetic.'' We conducted experiments using a dual-module architecture in which a reasoning module (an MLLM) interacts with an imagery module (Python/PyVista) on 3D model rotation tasks. Performance was lower than expected, with accuracy reaching at most 62.5\%. 
Further investigation suggests that even when the burden of maintaining and manipulating a holistic 3D state is outsourced, the system still fails. This reveals that current frontier models lack the foundational \emph{visual-spatial primitives} required to interface with imagery. Specifically, they lack: (1) the low-level sensitivity to extract spatial signals such as (a) depth, (b) motion, and (c) short-horizon dynamic prediction; and (2) the capacity to reason contemplatively over images, dynamically shifting visual focus and balancing imagery with symbolic and associative information.\\
Project page: \url{https://github.com/sergiohayashi/imagery--3d-rotation}
\end{abstract}

\newpage

% --- Table of Contents ---
\tableofcontents
\newpage

% --- Section 1: Introduction ---
\section{Introduction}
Human reasoning relies heavily on mental imagery \citep{shepard1971mental,kunda2018visual}---the ability to visualize and manipulate objects in the ``mind's eye'' \citep{kosslyn1994image}. Theories such as Paivio's Dual-Coding Theory \citep{paivio1986mental} and Kosslyn's Perceptual-Anticipation Theory \citep{kosslyn1980image} suggest that this capability is distinct from verbal reasoning. While Multimodal Large Language Models (MLLMs) process images, they treat them as sequences of tokens, lacking the holistic, transformable nature of human mental images \citep{vaswani2017attention, kosslyn1980image}.

This paper explores the hypothesis that current LLMs lack an internal mechanism for visual state simulation---a limitation we refer to as functional ``aphantasia''\footnote{We use `functional aphantasia' strictly as a metaphorical label for the operational deficits defined here; we make no claims about human cognition or clinical conditions.} (as an analogy for illustration purposes). While an external tool can theoretically supply the persistent 3D structural representation, our experiments show that LLMs still fail to utilize it. Concretely, we argue that the LLM side lacks the basic \emph{perceptual primitives} needed to close the reasoning loop. We operationalize these missing primitives as deficits in (1) extracting and tracking low-level spatial signals (e.g., depth, movement, short-horizon dynamic prediction), and (2) engaging in contemplative, attention-driven visual-spatial reasoning rather than defaulting to symbolic, word-based associations.

\section{The Human Mental Imagery System and Motivation}
The study of human mental models has a long history; these theories can help us understand the missing capabilities of LLMs when human cognition is used as a blueprint.

\subsection{Imagery Reasoning and Spatial Reasoning}

For over 2000 years, it was believed that the human mind functioned in a ``perceptual'' manner, based on mental images. However, between the 1950s and 1970s~\citep{fodor1975language}, influenced largely by linguists, the cognitive revolution took hold, and the prevailing view became that the mind is predominantly ``computational'' or symbolic---that is, it thinks via sequential symbol manipulation \citep{pylyshyn1973mind}.

Nevertheless, the view that human reasoning includes an imagery component persists. In this model, alongside sequential symbolic elements, there exists a representation of an ``abstract image,'' which possesses a structural, holistic, and hierarchical nature. This component accompanies, shapes, and supports the sequential processing (and is bidirectionally linked to it).

Recent evidence suggests that current LLMs exhibit behavior analogous to a verbal reasoning module, corresponding to the left hemisphere of the brain, while simultaneously displaying deficiencies in characteristics associated with the right hemisphere (spatial/holistic processing), thus renewing interest in examining LLMs through this lens.

According to Paivio's Dual Coding Theory \citep{paivio1986mental}, human cognition operates through two distinct systems: a verbal one, based on language (logogens), and a non-verbal one, based on mental images (imagens). \citet{kosslyn1980image} argues for the so-called Depictive Model of Imagery, which views images as closely related to visual perception. Barsalou's Perceptual Symbol Systems (PSS) posits that images can be shaped and simulated to support cognition \citep{Barsalou1999}. Similar theories have been proposed by \citet{Lake2017} and \citet{johnson1989mental}. Experiments such as those by \citet{shepard1971mental} helped reinforce the theory that people think visually, specifically by performing mental rotation of 3D images to solve matching problems (response times increase linearly with the angular disparity between the objects).

One characteristic of this model is role separation and the existence of bidirectional communication. Although the specific nature of this communication, in terms of exactly what information it carries, remains unclear, it is reasonable to assert that the imagery component provides visual information that supports reasoning. The imagery module is not static, but dynamic; it is shaped by the reasoning of the logogen and works collaboratively. It is not, for instance, merely a passive representation like an unchanging photograph.

Based on this premise and through this lens, we can draw a comparison with what is currently observed in LLMs. Recent studies have demonstrated interesting results, suggesting that imagery theory may provide indications on how to improve LLMs. We know that Transformer-based LLMs are inherently sequential and autoregressive, lacking an explicit, holistic structured representation useful for imagery, though this architectural trait alone does not inherently preclude visual reasoning capabilities.

However, recent studies have shown that strong parallels exist.

\citet{cai2025has} show that even the most recent models, with reasoning capabilities that beat human-level performance in many cognitive tests, fall significantly short in visual tasks that would be trivial for a human, such as 3D rotation. Another study \citep{lei2025large} shows that activation patterns resemble signals from the brain’s left hemisphere, although it also cites a slight resemblance to the right side.

A recent study \citep{wang2024picture} demonstrated that when a model is trained with images, it performs better even on problems that do not involve images (text-only), despite requiring text to solve the problems. In other words, images aid the model's overall encoding, demonstrating that the Transformer also benefits from training on images.

Some studies have tried to overcome these limitations in the context of the Transformer itself, maintaining its sequential, token-based nature, by forcing the inclusion of intermediate visual representations, which helps in solving problems of a visual nature. Attempts include using chain-of-thought prompting, encouraging the model to generate image tokens in many forms \citep{rose2023visual, li2025imagine, xu2025visual, yang2025machine}, or attempting to modify part of the model to generate images as latent representations in downstream layers \citep{yang2025machine}. Some approaches rely on external tools, for example, those that provide image or spatial data \citep{huang2025my}, feedback from external representations \citep{wang2025vgr}---which uses tags like \texttt{<sot>[x1,y1,x2,y2]<eot>}---or external world models \citep{ha2018world, wu2025spmem}. Others have tried to improve image representation by providing images at different scales \citep{shi2024we} or modifying the architecture slightly such that image attention is preserved when integrated with text models \citep{zhang2024wings}. \citet{huang2025my} implements the idea of providing visual information through an external perceiver agent, paired with a small purely textual LLM, thereby creating a multimodal model.

None of these models represents the imagery module as an independent module with bidirectional communication and image manipulation capabilities.

Other models have gone further. Instead of treating images as patches and tokens, as was done in the original Transformer integration, they represent them using differentiable elements, such as Gaussian Splatting \citep{kerbl20233d, shi2024language, wu2025video}. This facilitates the model's ability to manipulate data consistently, performing transformations, for example. While the model handles images more natively, it is difficult to believe that humans, when thinking with images, think in this way---with high-precision images and calculated 3D transformations. For humans, imagery in reasoning may more closely resemble a ``visual sketchpad'' \citep{BADDELEY197447}.

The study by \citet{su2025thinking} effectively summarizes the current consensus in this area, envisioning three progressive stages toward full imagery capability: (1) Tool-Driven Visual Exploration, (2) Programmatic Visual Manipulation, and (3) Intrinsic Visual Imagination. As the authors note, realizing this final stage requires a critical innovation: the architectural integration of generative and reasoning capabilities within a single, unified model. Our work aligns with stage (2) of this proposal.

\subsection{Object Imagery, Spatial Imagery, and Propositional Reasoning}

To better understand these visual reasoning dynamics, cognitive psychology often distinguishes between object imagery, spatial imagery, and propositional reasoning~\citep{jin2024systematic,kozhevnikov2005spatial}. Object imagery concerns visual qualities such as color, texture, shape, and fine pictorial detail. Spatial imagery concerns relations such as position, distance, movement, orientation, and transformation. Propositional reasoning, in contrast, relies on symbolic descriptions such as words, labels, or logical relations.

Under this taxonomy, standard LLMs excel at propositional reasoning but fundamentally struggle with spatial imagery. While multimodal models are increasingly capable of handling object imagery (e.g., identifying shapes or colors), they lack a native mechanism to perform the dynamic transformations—such as 3D rotations—inherent to spatial imagery. 

This distinction clarifies the terminology used in the rest of this paper. Our target phenomenon is not imagery in the broad sense, nor object imagery specifically, but spatial imagery: the ability to maintain and manipulate spatial structure through transformations. Similarly, what we previously described as the ``verbal'' component of an LLM is more precisely characterized as propositional reasoning.

\subsection{Motivation}

Based on these observations, we ask: what happens if we externally provide a tool that carries the burden of the hard work---namely, sustaining the spatial image representation and enabling manipulation---to support the propositional reasoning part? For this, we propose a compact system that incorporates an artificial imagery module and supports communication in a highly simplified setting. We then evaluate whether and to what extent this model improves performance; if it does not, we analyze the remaining limitations as steps toward genuine imagery-based reasoning.

\begin{figure}
    \centering
    \includegraphics[width=0.7\linewidth]{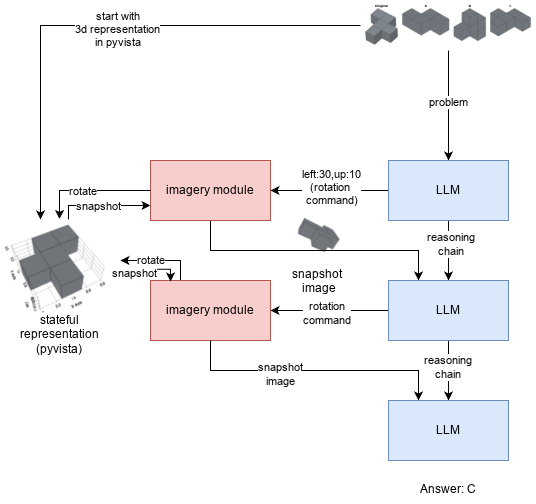}
    \caption{\textbf{Schematic of the visual feedback loop.} The LLM functions as the \textbf{Reasoning Module}, issuing rotation commands to the stateful \textbf{Imagery Module}. The imagery module processes these commands and returns a 2D snapshot of the object from the updated viewpoint.}
    \label{fig:imagery-loop}
\end{figure}

\section{Methodology: The Prosthetic Imagery Architecture}

To simulate the interaction between propositional reasoning and spatial imagery systems, we designed a dual-module agentic loop (Figure~\ref{fig:imagery-loop}) and evaluated it on the SpatialViz 3D rotation benchmark \citep{wang2025spatialviz}. The task is a standard mental rotation problem: given one \emph{original} object and three alternatives (A, B, C), the model must determine which alternative \emph{cannot} be obtained from the original by rotation alone \citep{shepard1971mental, wang2025spatialviz}.

This task is notoriously easy for humans, yet recent frontier models achieve very low scores, presenting a performance gap from human baselines of more than 45\% \citep{cai2025has}. The objects in the dataset are composed of uniform cubes, with no more than 2 levels in height and no more than 5 in width or depth (relative to axes). Thus, they have simple structures, and the views (rotated states) are such that they present no ambiguity or aliasing. Even a child can solve the problem without much difficulty, yet frontier models often fail.

\subsection{System Overview}

The system consists of two components.

\textbf{Reasoning Module.} A frontier multimodal LLM (primarily GPT-5.2 in the experiments reported here) acts as the agent that conducts the reasoning process and makes decisions. The model does not manipulate 3D geometry directly; it relies on the imagery module.

\textbf{Imagery Module.} This module is responsible for the imagery component of the process. It maintains a persistent 3D state for each object and executes manipulations requested by the reasoning module. It renders a 2D projection from the current camera angle, simulating human visual perception.

The problem-solving process is iterative: the reasoning module issues commands and ``sees'' the result from the updated angle. The imagery module is implemented in Python using PyVista. The flow is controlled by an external loop until the task is resolved.

\paragraph{Pose calibration / initialization.}
For each object in the problem, we performed a one-time manual camera-pose calibration to align the PyVista-rendered object with the viewpoint shown in the benchmark prompt. Concretely, we interactively adjusted rotation parameters (yaw/pitch/roll in camera space, i.e., relative to the current view) until the rendered snapshot visually matched the reference image. We then recorded the resulting rotation parameters and hard-coded them into the dataset construction script. During evaluation, the imagery module always starts from these fixed calibrated poses; no per-model or per-run adjustment is performed.

The rotation commands are defined in camera space as orbit-like movements around the object's center. The direction and magnitude are defined relative to the current state, using intuitive terms like left/right and up/down. For example, ``left:30'' rotates the object left by 30 degrees relative to the current view, analogous to turning it in your hands while your viewpoint remains fixed. There is no explicit axis definition, keeping it as close as possible to how humans think. We do not typically think in terms of $x$, $y$, and $z$ axes when rotating an object in our hands. Multiple commands for a single target are allowed, generating a grid of snapshots after each rotation step, which can provide a sense of continuous movement.

In each turn the reasoning module is provided with its memory (including the sequence of commands issued so far), the images generated in the last iteration, and the original problem statement. The system guarantees that all objects (whether as a result of a rotation command or in their current state) are provided at least once in each iteration.

The system is highly sensitive to prompt design. We arrived at the final prompt (Section~\ref{sec:prompts}) iteratively by observing failure modes and refining instructions until the LLM reliably engaged with the imagery module. This included explaining the iterative process, defining the command set, and establishing minimum incentives, such as requiring a minimum number of iterations.

Although prompting strategies can affect system performance, they cannot overcome the model's intrinsic limitations, nor do they qualitatively alter the core findings discussed in this paper.

\subsection{Model Selection}

Preliminary tests suggested GPT-5.2 performed best compared to other models. For reasons of cost, runtime, and reproducibility, our main experiments were conducted using GPT-5.2 (model GPT-5.2-2025-12-11).

We emphasize that the main qualitative behaviors observed in this paper were not unique to a single model. Across all frontier systems, we observed the same underlying pattern. Therefore, because our focus is more on qualitative analysis, we utilized GPT-5.2 as a representative baseline for current frontier MLLMs.

\subsection{Baseline}

According to the original SpatialViz study, the highest frontier-model performance on this task was 36.25\% (Claude-3.7-Sonnet), while human performance was approximately 79.16\% \citep{wang2025spatialviz}. More recent studies \citep{cai2025has} indicate that this gap persists even for newer models such as GPT-5 and Gemini-2.5-Pro. We also conducted a local, single-turn re-evaluation of current frontier models. As shown in Table~\ref{tab:spatialviz-scores}, GPT-5.2 (solving the problem directly without special prompting or tools) reaches 50.0\%, establishing a stronger baseline than previously reported systems, but still leaving a large gap to human spatial reasoning.

\section{Experiments and Results}

For ablation purposes and to highlight behavioral characteristics, we performed experiments under different conditions.

\subsection{Condition 1: Reset Enabled}

This setup serves as an ablation. Here, the model was allowed to use a reset operation to revert objects to canonical viewpoints. Because the 3D objects were constructed in exactly the same coordinate space, resetting two matching objects brings them to exactly the same canonical view.

Under this condition, the task is reduced to simple 2D image comparison, bypassing the need for dynamic rotation. In the best prompt variant, the model was explicitly encouraged to use the reset command (requesting a top view and horizontal view), and accuracy reached up to \textbf{97.5\%} (85--97.5\% across prompt variants), suggesting that identifying the odd one out was not fundamentally difficult and could be resolved in an unambiguous manner once perfectly aligned.

\subsection{Condition 2: Incremental Rotation with a 360$^\circ$ Search Hint}

Here, we instructed the model to trigger a sequence of rotations to obtain a full 360$^\circ$ trajectory view, providing examples of both short (6 steps) and long (20 steps) sequences. This approach theoretically allowed the model to simply search for a matching view and conclude immediately, bypassing the need to plan incremental rotations.

Although this strategy could bypass the need for spatial planning and fall back to pure image matching, surprisingly, the model achieved only 52.5\% accuracy (measured in a single run).

Despite making the task mechanically much easier, performance remained moderate. Qualitative analysis suggested that the model often struggled to focus on a specific matching frame when presented with a long grid sequence of snapshot images. Instead of identifying a matching frame directly, it frequently reverted to verbal inference about the object's structure. Furthermore, although a 360$^\circ$ orbit provides many facets, it is not guaranteed to hit the exact target angle; nonetheless, the primary bottleneck appeared to be the visual comparison process itself rather than the absence of corresponding viewpoints.

\subsection{Condition 3: Incremental Rotation Only}

This is our main experimental setup. The prompt was crafted to explain the mechanics of the iterative process, how to work with the tool, and the command definitions---without turning it into a step-by-step tutorial. We explicitly mentioned that using the tool makes it possible to reach the exact same angle as the target view (original). We also required at least 5 iterations, because otherwise the system tended to answer immediately relying solely on zero-shot reasoning. The full prompt is available in Appendix~\ref{sec:prompts}.

In this setup, we obtained a score ranging from \textbf{50\%} to \textbf{62.5\%} (across 10 runs, with the model and prompt frozen). An example of the resolution thread is presented in Appendix~\ref{sec:res-sample}.

Qualitative analysis showed that the model indeed used the imagery module (avoiding immediate answers), but treated it primarily as an aid for static object inspection. At no point did we observe the model actively confirming it had achieved a geometric match through continuous rotation. This behavior was consistent across all experiments and all evaluated models.

For illustration, the model's typical rationale resembled the following:

\begin{quote}
\itshape
\small
From rotations, original appears to be 5 cubes: a 2×2 top plate (4 cubes) with one extra cube attached below one corner (making a 'foot'). Option B visually matches this: 2x2 plate with a single cube sticking out from a side and one cube above? need verify carefully. Option A looks like a 1x4 bar with one cube on top (still 5) arranged differently (not 2x2). Option C is a 3-cube vertical column with a 2-cube horizontal arm at top and 1 cube at bottom side (total 5) but adjacency differs. Next: rotate original to see if can form A or B; also rotate B/A/C to check for 2x2 plate invariant. We'll do additional rotations focusing on revealing 2x2 face on each.
\end{quote}

\begin{figure}[htbp]
    \centering
    \includegraphics[width=0.6\textwidth]{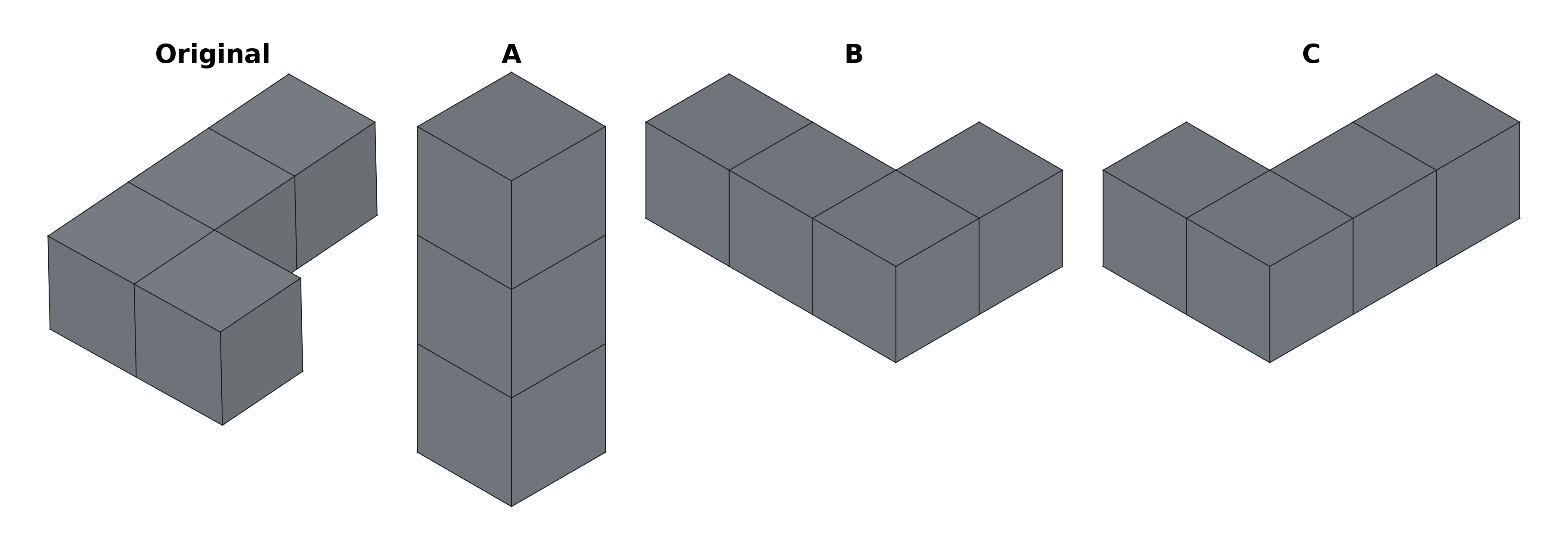}
    \caption{SpatialViz 3D Rotation sample problem. The question is: \textit{The left image shows the original cube stack made of equal-sized small cubes. Which of the options on the right cannot be obtained by rotating the original cube stack? Please answer from options A, B or C.}}
    \label{fig:spatialviz-3d-rotation-sample-problem}
\end{figure}

\begin{table}[h]
    \centering
    \caption{Evaluation of frontier models on the SpatialViz 3D rotation task (3DR), compared to human baselines. The bottom section details our experimental phases isolating spatial reasoning mechanisms, revealing that unrestricted manipulation yields artificially high performance (97.5\%) through 2D matching rather than true 3D spatial simulation.}
    \label{tab:spatialviz-scores}
    \begin{tabular}{@{}lc@{}}
        \toprule
        \textbf{SpatialViz - 3DR} \\
        \textbf{Evaluated Entity} & \textbf{Accuracy (\%)} \\ \midrule
        \textit{Human Performance}\citep{wang2025spatialviz} & \textit{79.16} \\ \midrule
        \multicolumn{2}{@{}l}{\textbf{Baseline Models}} \\
        Llama-4-Maverick-17B-128E-Instruct\citep{wang2025spatialviz} & 40.00 \\
        Claude-3.7-Sonnet\citep{wang2025spatialviz} & 36.25 \\ \midrule
        \multicolumn{2}{@{}l}{\textbf{Baseline Models - Our Re-evaluation}} \\
        GPT-5.2 & 50.00 \\
        Gemini-3.5-Pro & 40.00 \\
        Claude-3.7-Sonnet (20250219) & 37.50 \\
        Claude-Opus-4.5 (20251101) & 37.50 \\
        Grok-4.1-Fast-Reasoning & 37.50 \\
        Qwen3-VL-30B-A3B-Instruct & 20.00 \\ \midrule
        \multicolumn{2}{@{}l}{\textbf{Experiment results}} \\
        Ablation condition 1: With reset (canonical views) & \textbf{85}--\textbf{97.5} \\
        Ablation condition 2: With 360$^{\circ}$ view hint & \textbf{52.5} \\
        Normal condition: Incremental rotation, minimal prompt & \textbf{55}--\textbf{62.5} \\
        \bottomrule
    \end{tabular}
\end{table}

\subsection{Prompt Invariance and Cognitive Limits}
LLMs are highly sensitive to prompting. We conducted extensive prompt engineering to rule out deficiencies stemming solely from poor prompts, although we did not attempt to find the absolute maximum possible accuracy through overfitting. Variations tested included:
\begin{itemize}
    \item Setting a minimum number of iterations (otherwise, the model answers immediately without relying on rotations).
    \item Instructing the model to use the imagery module to perform continuous rotations and attempt a direct geometric match to the target, prioritizing this over structural deduction.
    \item Step-by-step resolution: first grasping the overall structure, then focusing on specific aspects or alternative objects.
    \item Tracing the partial solution by classifying each alternative (e.g., 'solution candidate', 'no match').
    \item Estimating the rotational distance before emitting rotation instructions.
\end{itemize}

None of these variations yielded meaningful improvements beyond the 62.5\% ceiling.

\subsection{Model ablation study}
\label{subsec:model-ablation}

We conducted an ablation study across different frontier LLMs under strictly identical conditions. \texttt{gpt-5.2} demonstrated the most reliable tool-use behavior, leading us to optimize and freeze our prompt design based on its responses, achieving 50.0\%--62.5\% in the final experiments. Interestingly, \texttt{gpt-5.4} performed slightly worse under the same conditions, yielding 45.0\% to 52.0\% across 3 runs. Other models struggled significantly to maintain the rigorous, multi-step agentic loop required for the task: \texttt{gemini-3-pro-preview} failed to complete the benchmark even after 12 hours of execution, \texttt{gemini-3-flash-preview} suffered from severe hallucinations forcing us to halt execution, and \texttt{grok-4.1-fast-reasoning} could not follow the output syntax (e.g., outputting \texttt{cw} instead of the required \texttt{rotate:cw}, or simply ``0''). Consequently, we utilized \texttt{gpt-5.2} as our primary experimental vehicle due to its unmatched operational stability in this specific tool-use environment.

\subsection{Isolating Core Spatial Capabilities}

\subsubsection{Identify small rotation}
To understand the lower-than-expected performance, we isolated the fundamental unit of spatial reasoning: detecting rotation. We presented the models (GPT-5.2, Gemini 3 Pro) with two images of the same object (from the SpatialViz dataset), where the second was rotated by a small angle in a specific direction (yaw, pitch, or roll). We then asked: ``Estimate the rotation direction and angle.''

\begin{figure}
    \centering
    \includegraphics[width=0.5\linewidth]{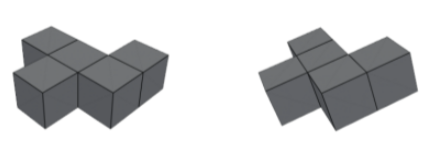}
    \caption{The models were asked to detect the rotation (direction and angle) required to transform the left image into the right image. The correct answer is ``left:30''. All tested models answered incorrectly: GPT-5.2 predicted ``right:90'', GPT-5.1 predicted ``right:45'', and Gemini-3-Flash predicted ``rotate:ccw:35,left:45,up:20''. Here, ``ccw'' means counterclockwise rotation.}
    \label{fig:rotate-problem}
\end{figure}

In a test with three rotation axes and the rotation magnitude fixed at $15^\circ$, the results for GPT-5.2 are summarized in Table~\ref{tab:rotation-results}. It correctly predicted only one condition (Right $15^\circ$).

The failure modes reveal a profound inability to infer motion from two sequential images. For instance, when the object was rotated \emph{left}, the model predicted \emph{right}. Crucially, in the \emph{down} rotation scenario, the model incorrectly predicted \emph{up}, justifying its answer by noting that ``the top part is visible''---conflating static feature visibility with the direction of motion. Similarly, planar rotations (clockwise and counterclockwise) yielded unrelated directional predictions (e.g., predicting ``right'' or ``up''). A more detailed explanation of how the rotation was generated (in camera space) did not improve these results.

\begin{figure}
    \centering
    \includegraphics[width=0.5\linewidth]{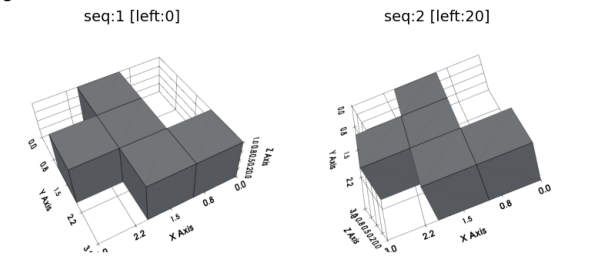}
    \caption{VGGT correctly recognized the rotation direction and angle.}
    
    \begin{quote}
    \small
    Translation (Direction): [-0.3297236  -0.01104623  0.07304744] \\
    Rotation (Degrees): [-0.90105244 18.20889242  1.1002316 ] \\
    Pitch (X): -0.90$^\circ$ \\
    Yaw (Y): 18.21$^\circ$ \\
    Roll (Z): 1.10$^\circ$
    \end{quote}
        
    \label{fig:vggt}
\end{figure}

Regarding this specific capability, certain specialized vision transformers, such as VGGT ~\citep{wang2025vggt}, are explicitly trained to recover camera extrinsics. In the example shown in Figure~\ref{fig:rotate-problem}, VGGT correctly output a yaw of 18.21$^{\circ}$ (within a margin of error of less than 2 degrees). See Figure~\ref{fig:vggt}.

\subsubsection{VGGT camera-transformation probe}

To systematically evaluate VGGT's robustness, we generated a controlled dataset using the SpatialViz 3D models by applying incremental camera orbits in three intuitive directions---yaw (left/right), pitch (up/down), and roll (clockwise)---from 0$^{\circ}$ to 360$^{\circ}$ in 30$^{\circ}$ steps. VGGT's prediction correctness was verified by re-applying the inferred rotations (Euler angles) in PyVista and manually checking visual alignment with ground-truth snapshots. Results showed perfect matching for all yaw angles and accurate predictions for pitch up to $\pm 90^{\circ}$ (and 330$^{\circ}$), but it failed on extreme pitch angles (120$^{\circ}$--270$^{\circ}$) and most roll angles beyond $\pm 30^{\circ}$. These failures manifested as axis-reversed interpretations due to the model's training bias toward upright real-world poses. Manual validation was straightforward, as the match was either visually perfect or mirrored across an axis (which we considered a failure).

\subsubsection{Single-object rotation probe}

We further evaluated VGGT on the full 3D rotation dataset, consisting of 40 problems and 80 image pairs. Each pair involved transforming the ``original'' view to one of the matchable alternatives, specifically two out of the three candidate views of the original problem. In this setting, VGGT succeeded in 41 out of 80 cases, corresponding to an accuracy of 51.25\%. Given the two images, VGGT predicted a direct transformation, parameterized by Euler angles. Correctness was verified by applying the predicted transformation and visually inspecting the result.

For comparison, the MLLM with the Imagery Module was asked to iteratively rotate the candidate image until it matched the target, with a maximum of 12 iterations. Under this setting, the model aligned the images correctly in only 14 out of 80 cases, corresponding to an accuracy of 17.5\%.

Considering that VGGT is a Transformer-based model, this result provides strong evidence that the Transformer architecture is not inherently a bottleneck for learning geometric extraction. The limitations observed in current generalist MLLMs likely stem from training strategies rather than architecture; specifically, generalist models lack the supervised training on camera extrinsics found in datasets like Co3D. However, we must also note that MLLMs are primarily text-centric reasoning systems, operating very differently from specialized models like VGGT.

\begin{figure}
    \centering
    \includegraphics[width=0.5\linewidth]{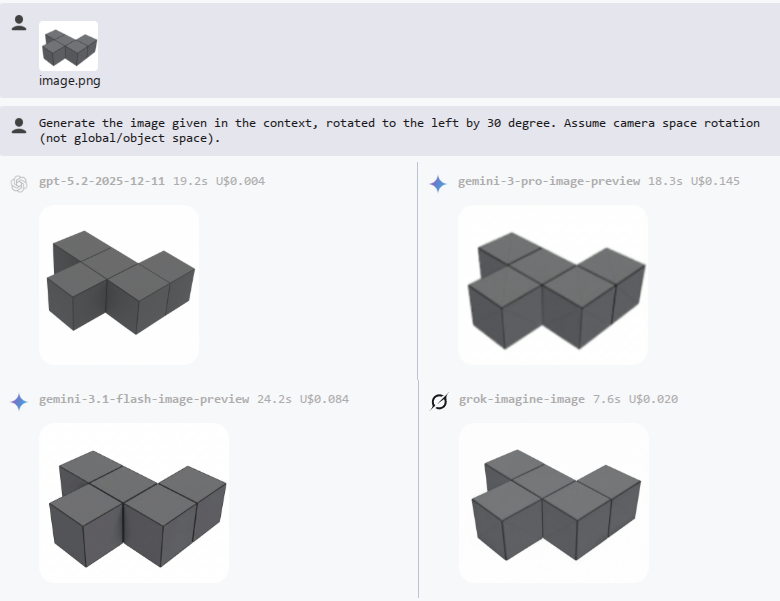}
    \caption{The models were asked to generate an image rotated by 30 degrees to the left (camera space). None of the models were able to do so correctly, or even produce a close approximation.}
    \label{fig:rotate-image-models}
\end{figure}

\subsubsection{Predict the after-rotation state}

Beyond the capability to perceive rotation (identifying direction and angle), spatial reasoning requires the model to conceptually visualize the resulting state when a rotation is applied---for example, simulating a small rotation in an arbitrary direction in order to issue the correct rotation instruction. To investigate this capability, we asked four frontier image-generation models (\texttt{gpt-5.2}, \texttt{gemini-3-pro-image-preview}, \texttt{gemini-3.1-flash-image-preview}, \texttt{grok-imagine-image}) to generate a post-rotation image given an initial object image. All models failed completely (Figure~\ref{fig:rotate-image-models}); they simply output the original image unchanged. Prompt variations did not alter this outcome.

\begin{table}[h]
    \centering
    \caption{Performance of GPT-5.2 on 15-degree rotations across different axes. The model fails to generalize directionality, relying instead on spurious visual cues.}
    \label{tab:rotation-results}
    \begin{tabular}{@{}lllc@{}}
        \toprule
        \textbf{True Direction} & \textbf{Angle} & \textbf{Model Prediction} & \textbf{Outcome} \\ \midrule
        Left & $15^\circ$ & Right & \ding{55} \\
        Right & $15^\circ$ & Right & \ding{51} \\
        Up & $15^\circ$ & Right ($30^\circ$) & \ding{55} \\
        Down & $15^\circ$ & Up (reason: top visible) & \ding{55} \\
        Clockwise & $15^\circ$ & Right ($30^\circ$) & \ding{55} \\
        Counter-Clockwise & $15^\circ$ & Up & \ding{55} \\ \bottomrule
    \end{tabular}
\end{table}

\section{Discussion}

\subsection{Functional spatial-imagery deficit (metaphor: `functional aphantasia')}
We use ``functional aphantasia'' strictly as a conceptual metaphor to encapsulate several measurable computational deficits observed in our experiments. Even when equipped with an external tool that performs spatial rotations and prompted to engage in imagery-based thinking, the model remains unable to effectively ``simulate'' rotation to solve the problem. Crucially, our architecture successfully outsources the maintenance of the persistent 3D model to the programmatic module. Therefore, the failure of the system highlights that holding a 3D state is only half the battle. To actually use this state, the LLM must possess basic visual-spatial primitives to interpret the generated snapshots. We identify the lack of these foundational primitives as the core cause of the imagery deficit:

\textbf{Insensitivity to movement:} When provided with two images, the model consistently fails to perceive them as a sequence of snapshots of an object in motion. It treats them as isolated, unrelated frames, even when explicitly prompted to view them as a human naturally would---recognizing the same object and ``imagining'' its continuous movement.

The results also suggest that the model lacks the computational equivalent of ``stereo matching''~\citep{yang2025machine} to extract spatial information within and between images. In humans, binocular vision provides depth cues that make 3D structure self-evident, allowing us to solve spatial matching problems by mentally simulating rotation step by step. LLMs, lacking this spatial intuition, treat rotation primarily as abrupt 2D changes. Specialized solutions like VGGT~\citep{wang2025vggt} show promise---achieving $\sim$51\% accuracy on our 3D dataset with near-perfect yaw recovery---though they remain limited by dataset biases regarding large pitch and roll angles. Consequently, these findings suggest that a more realistic path forward is to explicitly model spatial reasoning via pose tracking and projection, which our results suggest is currently missing in general-purpose LLMs.

\textbf{Inability to predict dynamic states:} The model cannot ``imagine'' or generate the state of an object after a rotation, as demonstrated by our simple experiment with frontier image models (Figure~\ref{fig:rotate-image-models}). In principle, an MLLM could employ a generate-and-test strategy---synthesizing intermediate imagery and iteratively analyzing the results---but this is currently hindered by severely limited reliability in visual-spatial prediction.

\textbf{Symbolic-associative reasoning bias:} Previous research~\citep{hayashi2022understanding} suggests that autoregressive models exhibit a strong bias toward textual sequences, driven by the higher predictability of text compared to non-sequential image tokens. As a result, these models default to symbolic inference, often disregarding nuanced visual cues. The model-generated rationales in our experiments indicate that MLLMs do not attend to images in a focused, human-like manner; instead, they reason by grasping a superficial gist of an image and advancing their logic purely through symbolic links. For instance, the model struggles to deliberately inspect a specific region of an image or to ``contemplate'' (i.e., dynamically hover its attention over) visual features, which is characteristic of human visual engagement. Its reasoning is ultimately grounded in shallow perceptual observations rather than a robust structural understanding.

In our experiments, this bias proved robust against intervention: even with explicit prompting designed to force visual-spatial matching, the model persistently reverted to propositional reasoning about object composition rather than engaging with visual geometric features.

\section{Conclusion}
This study provides evidence that current frontier LLMs exhibit what we metaphorically call ``functional aphantasia,'' more precisely a functional spatial-imagery deficit: they do not reliably engage in spatial-imagery-like reasoning even when provided with an external tool capable of rendering and rotating 3D objects.

Across our experiments, the external Imagery Module functions as a capable ``scanner'' (generating relevant views on command), but it remains an insufficient ``cognitive prosthetic'' when paired with today’s LLM Reasoning Modules. By providing the persistent 3D representation externally, we isolated what is truly missing on the LLM side: the basic perceptual primitives. The core limitation is not the absence of a rendered 3D state, but the model’s fundamental inability to process the visual-spatial primitives required to interface with that state.

In this study, we identified two central barriers. First, the model lacks the foundational primitives for spatial and visual processing, specifically a severe insensitivity to (i) \textbf{depth} (extracting 3D structure from 2D projections), (ii) \textbf{movement} (interpreting two images as temporally related snapshots), and (iii) \textbf{dynamic prediction} (anticipating the subsequent view after a geometric transformation). Without these building blocks, the LLM cannot guide the imagery module. Second, the model lacks a \textbf{contemplative, attention-driven} mode of reasoning over images; instead, it defaults to symbolic, word-based associations that frequently override spatial evidence.

Accordingly, future work should move beyond merely attaching LLMs to rendering engines or adding parallel imagery branches. Instead, research must focus on building foundational architectures that intrinsically support spatial awareness, continuous state-tracking, and contemplative visual attention.

\newpage

% --- References ---
\bibliographystyle{apalike}
\bibliography{references}

\newpage
% --- Appendix ---
\appendix
\section{Prompts}
This is the freeze version of the prompt.

\label{sec:prompts}
\begin{lstlisting}[basicstyle=\ttfamily\scriptsize, breaklines=true]

# TASK AND ITERATIVE PROCESS
Your task is to solve a 3D model rotation problem.

The problem includes an image with 4 figures, and the following statement:
`The left image shows the original cube stack made of equal-sized small cubes. Which of the options on the right cannot be obtained by rotating the original cube stack? Please answer from options A, B or C.`

# IMAGE MODULE
To solve this problem, you will work together with a tool called the imagery module.

The imagery module holds a 3D representation of the problem objects and perform rotation operations on your behalf, and generate snapshots (images) corresponding to the current state (i.e., camera angle). The state of each object is maintained throughout the entire process. The initial state (camera angle) of each object corresponds to the image in the problem statement.

The problem asks whether one object can have the same view as the other through rotation. The imagery module helps solve the problem by actually performing the rotation and providing the view after-rotation, enabling a try-rotate and check loop process, you don't need to "imagine" it. You can request a direct rotation to a desired final target state or do it incrementally, in a loop rotate-verify until get the desired view of conclude that is not possible. It is like take the objects in you hands, and play if around checking visually if you have a match.

The problem asks to rotate the original to match the alternative, but for the problems presented here, it is equivalent rotate the alternative to match the original. The imagery module allows rotate only the alternatives.

Working with the imagery module is an iterative process, controlled externally. It works in TURNS between you and the imagery module. On your turn, do the analysis based on the provided images, and generate rotation instructions to the imagery module. Then, the imagery module, on its turn, will apply these rotations and return the snapshot images of the objects in the new state. Then it is your turn, and so on.

Rotations commands are defined in camera space (relative to the current view), simulating the inverse of camera movement. Intuitively, this matches the view of manipulating an object in your hands: the object spins around its center while the camera (your viewpoint) remains fixed. 

Possible commands are:
- `left:value` (object is rotated to left)
- `right:value` (object is rotated to right)
- `up:value` (object is rotated up)
- `down:value` (object is rotated down)
- `rotate:cw:value` (object rotates clockwise in the image plane)
- `rotate:ccw:value` (object rotates counterclockwise in the image plane)

`value` refers to the rotation angle in degrees. Angle 0 is also valid and can be used to get a snapshot of the current state.

# OUTPUT
Return your response in JSON format, following the format below:
```json
{
  "memory": {
    "rationale": "your justification up to this point",
    "partial_conclusion": {
      "A": "unknown"|"probably_not_the_answer"|"probably_the_odd_one",
      "B": "unknown"|"probably_not_the_answer"|"probably_the_odd_one",
      "C": "unknown"|"probably_not_the_answer"|"probably_the_odd_one"
    }
  },
  "iteration_number": 1,
  "commands": [
    {
      "target": "A"|"B"|"C",
      "rotation_sequence": "right:15,right:15,up:10"
    }
  ],
  "final_answer": null
}
```

Details of the output fields:
- `memory`: Generate your rationale and partial conclusion to help trace your reasoning process. This block will be provided as context in future turns during the iteration, so it will serve as your memory throughout the iterative process.
- `commands`: Rotation instructions for the imagery module. You can generate for one or more targets. Rotation sequence can have one or more commands, separated by comma. Each command generates a snapshot image of after rotation view, and will be combined in a grid image, per target, having the effect of a sequence showing the object rotating incrementally. 
- `final_answer`: The answer for the problem, if you have a conclusion. Otherwise, leave as null.
- `iteration_number`: Iteration counter. Start with 1 and increment this number each turn.

Enclose the JSON object in ```json and ```.

*IMPORTANT*
In you turn, generate exactly one JSON output and FINISH. DON'T simulate the iteration or the imagery module turn. It is handled externally.

# CONVERSATION CONTEXT
The conversation context, in each turn, will contain the following content:
- The text and image from the problem statement.
- All the previous output you have generated.
- The images generated by the imagery module, from the last iteration only.
- The `original` object snapshot to help comparison.

# STRATEGY
- Perform at least 5 iterations before giving the final answer.
\end{lstlisting}

\section{Resolution sample}
\label{sec:res-sample}

\noindent
\textbf{Question:}

\textit{The left image shows the original cube stack made of equal-sized small cubes. Which of the options on the right cannot be obtained by rotating the original cube stack? Please answer from options A, B or C.
}

\medskip
\hrule height 0.1pt
\medskip

\noindent
\begin{minipage}{0.48\textwidth}
\centering
\includegraphics[width=0.9\textwidth, height=2cm, keepaspectratio]{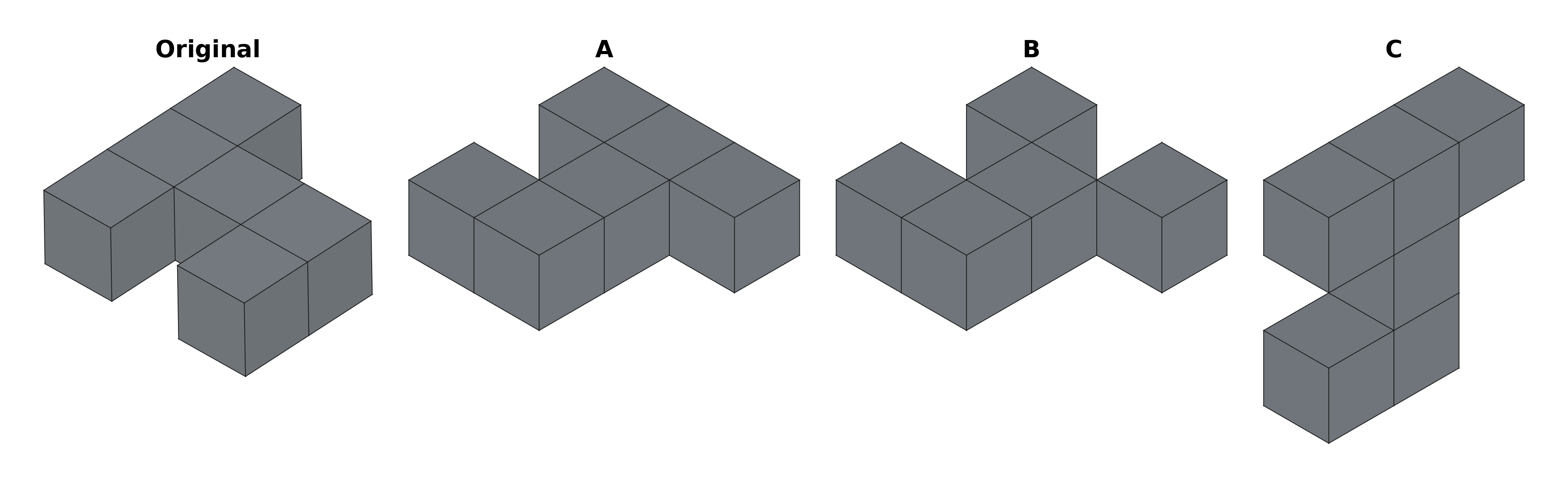} \\[0.5ex]
\end{minipage}
\hfill
\begin{minipage}{0.48\textwidth}
\textbf{Iteration 1:} \\
\footnotesize{
Need determine which option shape is not a rotation of original. Use imagery module to rotate each option to see if can match original view. Start by checking overall cube count/adjacency: original appears to be 5 cubes: a 2x2 platform of 4 cubes with one extra cube attached below/front? plus one cube stacked on back-right? Options show similar counts but unclear; must verify via rotations. First iteration: request snapshots of A,B,C without rotation (0 deg) to compare, then begin rotating likely candidates.}
\end{minipage}
\medskip 
\hrule height 0.1pt
\medskip 

\noindent
\begin{minipage}{0.48\textwidth}
\centering
\includegraphics[width=0.9\textwidth, height=2cm, keepaspectratio]{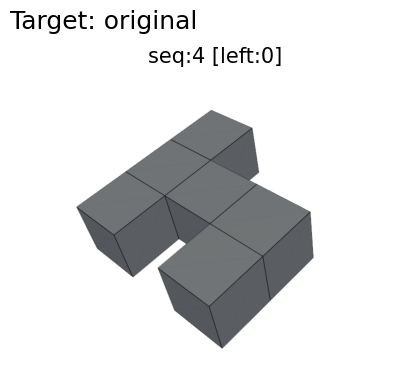} \\[0.5ex]
\includegraphics[width=0.9\textwidth, height=2cm, keepaspectratio]{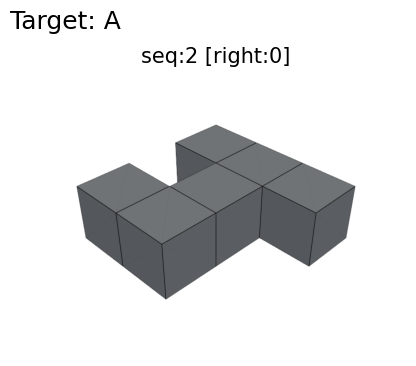} \\[0.5ex]
\includegraphics[width=0.9\textwidth, height=2cm, keepaspectratio]{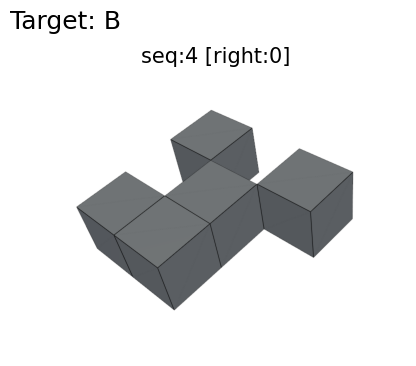} \\[0.5ex]
\includegraphics[width=0.9\textwidth, height=2cm, keepaspectratio]{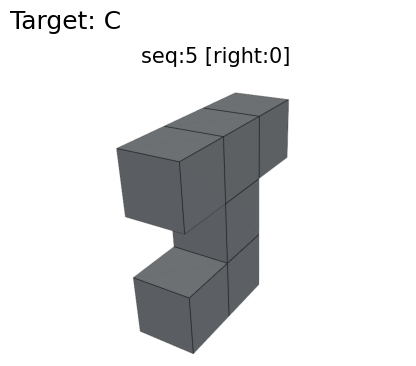} \\[0.5ex]
\end{minipage}
\hfill
\begin{minipage}{0.48\textwidth}
\textbf{Iteration 2:} \\
\footnotesize{
From snapshots: Original looks like 6 cubes: a 2x2 flat plate (4 cubes) plus one extra cube extending out on one side (making a 3-long arm) and one cube stacked on the far corner (height 2 at one corner). Option A appears to be a 5-cube shape (a 4-long bar with one cube stacked on second position) or possibly 6 but arrangement seems different; need rotate to verify adjacency. Option B looks closer to original: a 2x2 base with one cube stacked and one cube protruding. Option C is a vertical column of 3 with a 3-long top arm and one bottom cube offset—seems like 6 cubes but very different; likely not a rotation if original has only one cube at height 2 and no height-3 column. Need rotations to confirm matches: try rotate B to match original first; also rotate A and C to see if any can match original silhouette.}
\end{minipage}
\medskip
\hrule height 0.1pt
\medskip

\noindent
\begin{minipage}{0.48\textwidth}
\centering
\includegraphics[width=0.9\textwidth, height=2cm, keepaspectratio]{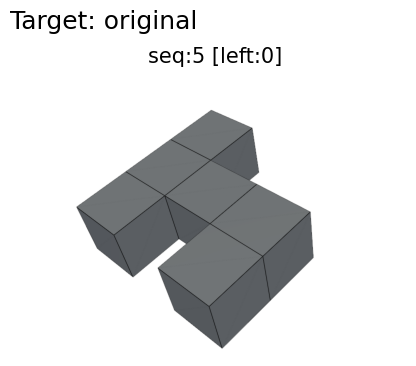} \\[0.5ex]
\includegraphics[width=0.9\textwidth, height=2cm, keepaspectratio]{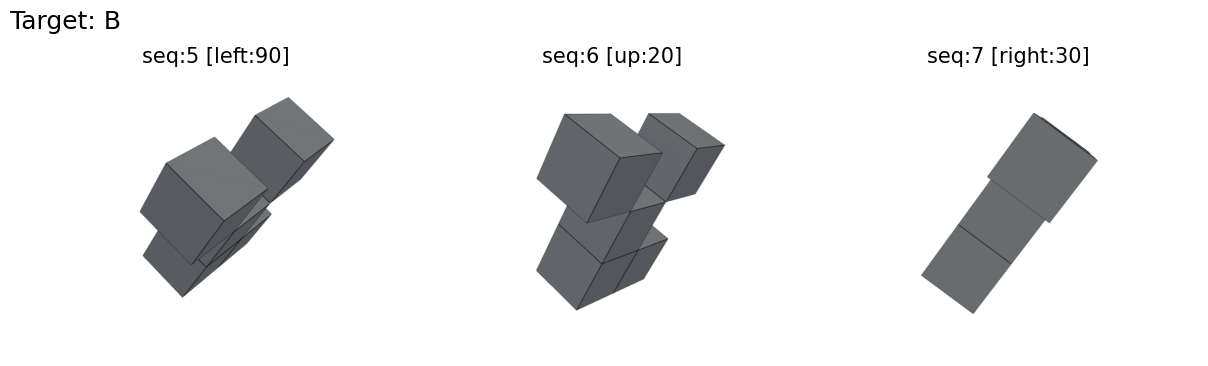} \\[0.5ex]
\includegraphics[width=0.9\textwidth, height=2cm, keepaspectratio]{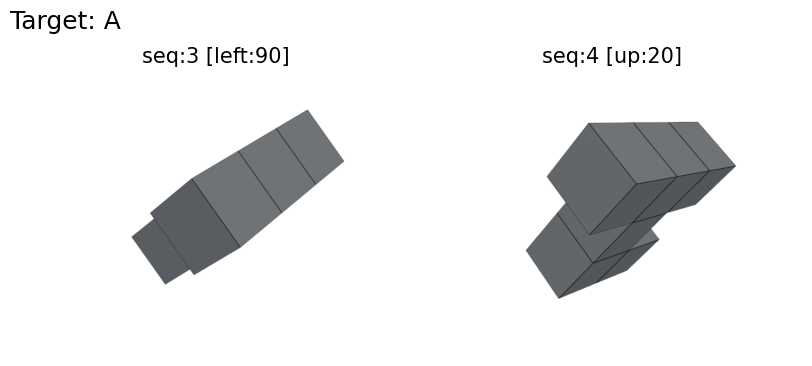} \\[0.5ex]
\includegraphics[width=0.9\textwidth, height=2cm, keepaspectratio]{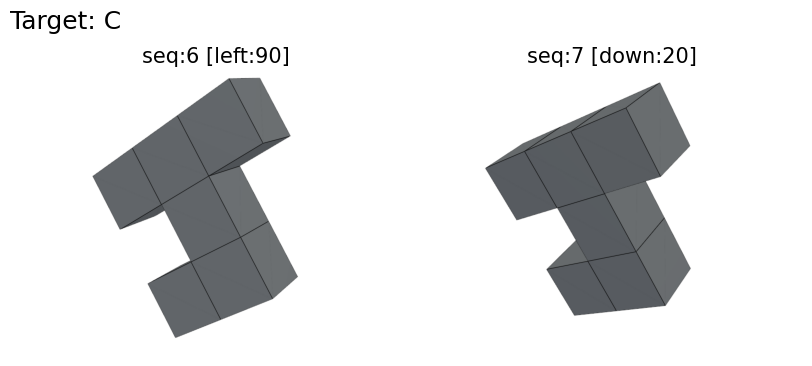} \\[0.5ex]
\end{minipage}
\hfill
\begin{minipage}{0.48\textwidth}
\textbf{Iteration 3:} \\
\footnotesize{
After rotations, C clearly has a 3-cube vertical column (height 3) with a 3-long top bar, which original does not (original max height seems 2). So C is very likely impossible. A and B rotations produced confusing edge-on views; need more systematic: try to match original's footprint: looks like a 2x2 square with one extra cube extending from one side (making a 3-long row) and one cube stacked on the far corner of the 2x2 (height 2). We'll attempt to rotate B to show a 2x2 base + one protrusion + one stacked. Also rotate A to see if it can ever show 2x2 base; currently A seems like a long bar with a step, maybe not matching either. Next: rotate B around to get clearer top-down-ish view; rotate A similarly; keep C but likely odd}
\end{minipage}
\medskip
\hrule height 0.1pt
\medskip

\noindent
\begin{minipage}{0.48\textwidth}
\centering
\includegraphics[width=0.9\textwidth, height=2cm, keepaspectratio]{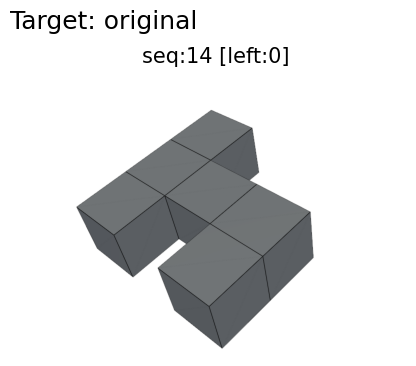} \\[0.5ex]
\includegraphics[width=0.9\textwidth, height=2cm, keepaspectratio]{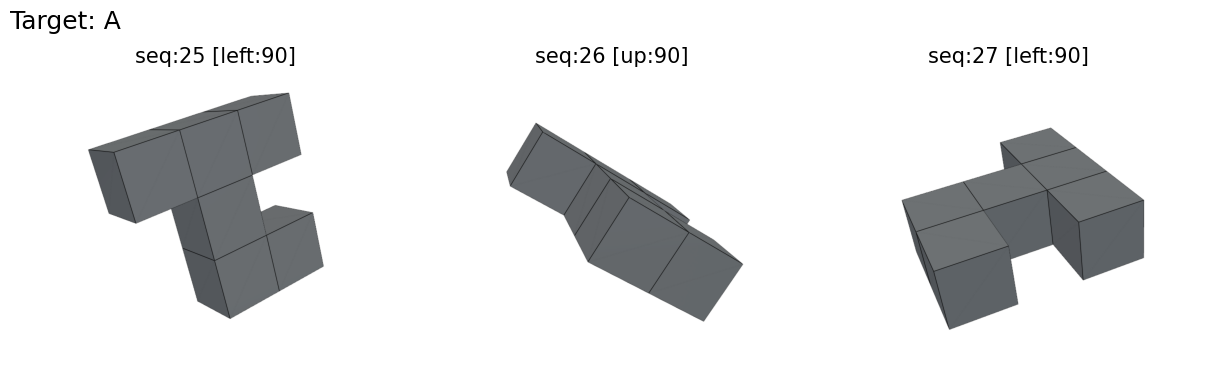} \\[0.5ex]
\includegraphics[width=0.9\textwidth, height=2cm, keepaspectratio]{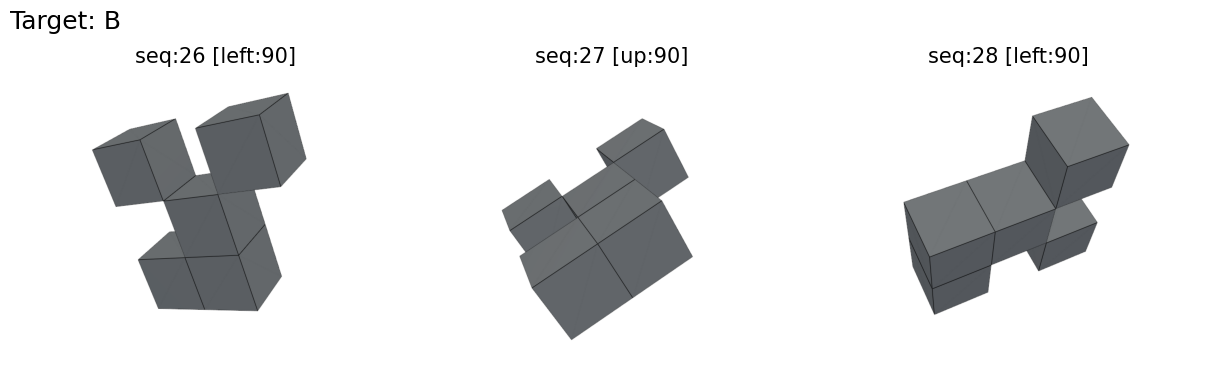} \\[0.5ex]
\end{minipage}
\hfill
\begin{minipage}{0.48\textwidth}
\textbf{Iteration 12:} \\
\footnotesize{
Latest snapshots show original is a flat (all cubes coplanar) 6-cube polycube. Option A after rotations (seq27) becomes a flat 6-cube footprint that visually matches the original’s ‘4-long bar with a 2-cube perpendicular arm’ arrangement, so A is obtainable. Option B in all shown views remains non-coplanar (has cubes at two heights: a cube stacked above the bar and another below), which cannot result from rotating a coplanar original (rigid rotations preserve coplanarity). Therefore B is the only option that cannot be obtained by rotation.}
\end{minipage}
\medskip
\hrule height 0.1pt
\medskip

\noindent
\begin{minipage}{0.48\textwidth}
\centering
\includegraphics[width=0.9\textwidth, height=2cm, keepaspectratio]{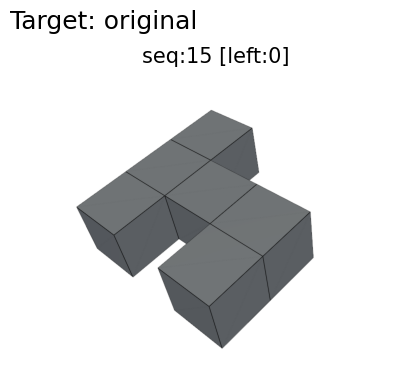} \\[0.5ex]
\includegraphics[width=0.9\textwidth, height=2cm, keepaspectratio]{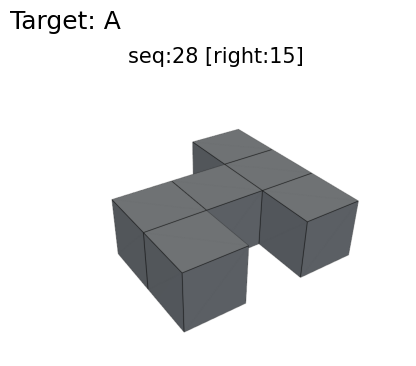} \\[0.5ex]
\includegraphics[width=0.9\textwidth, height=2cm, keepaspectratio]{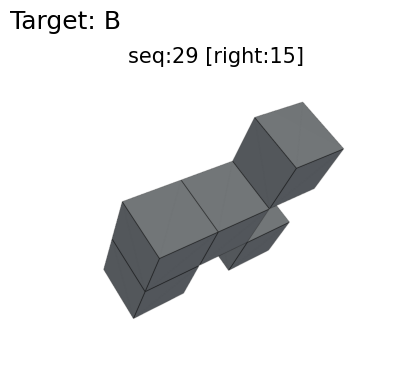} \\[0.5ex]
\includegraphics[width=0.9\textwidth, height=2cm, keepaspectratio]{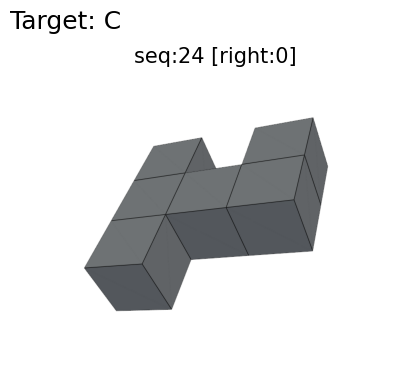} \\[0.5ex]
\end{minipage}
\hfill
\begin{minipage}{0.48\textwidth}
\textbf{Iteration 13} \\
\footnotesize{Comparing last snapshots: Original is a flat (single-layer) 6-cube polycube with a 4-cube bar and a 2-cube perpendicular arm. Option A is also flat and matches this footprint under rotation. Option C is flat too and could be another rotation of the same footprint. Option B is clearly non-coplanar (has cubes at different heights / stacked), which cannot be produced by rotating a coplanar original because rigid rotations preserve coplanarity. Therefore B cannot be obtained by rotating the original.} \\
Answer: \textbf{B}
\end{minipage}
\vspace{1em}

In this sample (which is representative of all observed cases), the model never demonstrates controlled rotation to bring the silhouette closer to the original. It fails to gauge angular distance---even when very close---often rotating in the opposite direction and moving further away. The chosen rotations frequently appear arbitrary, highlighting an inability to predict the post-rotation state. Furthermore, the generated rationale is almost entirely devoted to inferring the static structure of the object rather than planning spatial transformations.

\end{document}